\documentclass[letterpaper]{article} 
\usepackage{aaai24}  
\usepackage{times}  
\usepackage{helvet}  
\usepackage{courier}  
\usepackage[hyphens]{url}  
\usepackage{graphicx} 
\usepackage{natbib}  
\usepackage{caption} 
\usepackage{algorithm}
\usepackage{algorithmic}
\usepackage[flushleft]{threeparttable}
\usepackage[table]{xcolor} 
\usepackage{multirow}
\usepackage{amsmath}
\usepackage{amssymb}
\usepackage{ragged2e}
\usepackage{xspace}
\usepackage{makecell} 
\usepackage{booktabs}
\usepackage{array}
\usepackage{pifont}
\newcommand{\cmark}{\ding{51}}%
\newcommand{\STAB}[1]{\begin{tabular}{@{}c@{}}#1\end{tabular}}

\usepackage{algorithm}
\usepackage{algorithmic}

\usepackage{newfloat}
\usepackage{listings}
\DeclareCaptionStyle{ruled}{labelfont=normalfont,labelsep=colon,strut=off} 
\floatstyle{ruled}
\newfloat{listing}{tb}{lst}{}
\floatname{listing}{Listing}
\title{Modeling Continuous Motion for 3D Point Cloud Object Tracking}
\author {
    Zhipeng Luo\textsuperscript{\rm 1,\rm 2}\footnote{Work done at S-Lab, Nanyang Technological University},
    Gongjie Zhang\textsuperscript{\rm 1},
    Changqing Zhou\textsuperscript{\rm 3},\\
    Zhonghua Wu\textsuperscript{\rm 3},
    Qingyi Tao\textsuperscript{\rm 3},
    Lewei Lu\textsuperscript{\rm 3},
    Shijian Lu\textsuperscript{\rm 1}\footnote{Corresponding author}
}
\affiliations {
    \textsuperscript{\rm 1}S-Lab, Nanyang Technological University\\
    \textsuperscript{\rm 2}Black Sesame Technologies\\
    \textsuperscript{\rm 3}SenseTime Research\\
    zhipeng001@e.ntu.edu.sg, shijian.lu@ntu.edu.sg
}

\begin{document}

\maketitle

\begin{abstract}
The task of 3D single object tracking (SOT) with LiDAR point clouds is crucial for various applications, such as autonomous driving and robotics. However, existing approaches have primarily relied on appearance matching or motion modeling within only two successive frames, thereby overlooking the long-range continuous motion property of objects in 3D space. To address this issue, this paper presents a novel approach that views each tracklet as a continuous stream: at each timestamp, only the current frame is fed into the network to interact with multi-frame historical features stored in a memory bank, enabling efficient exploitation of sequential information. To achieve effective cross-frame message passing, a hybrid attention mechanism is designed to account for both long-range relation modeling and local geometric feature extraction. Furthermore, to enhance the utilization of multi-frame features for robust tracking, a contrastive sequence enhancement strategy is proposed, which uses ground truth tracklets to augment training sequences and promote discrimination against false positives in a contrastive manner. Extensive experiments demonstrate that the proposed method outperforms the state-of-the-art method by significant margins on multiple benchmarks.
\end{abstract}

\begin{figure}[t!]
    \centering
    \includegraphics[width=0.9\linewidth]{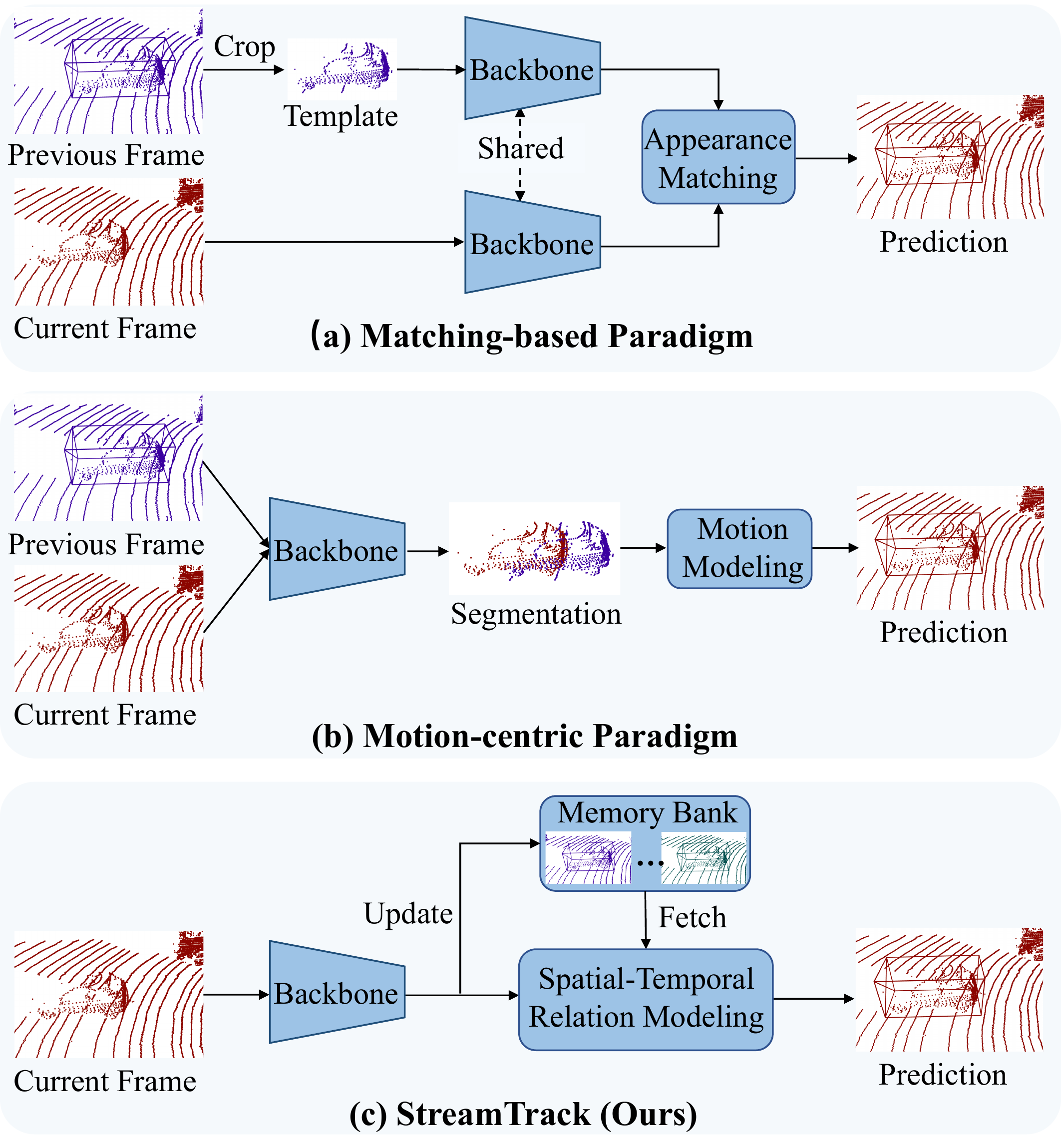}
    \caption{Comparison of 3D single object tracking paradigms. (a) The matching-based paradigm extracts features from a cropped template and a search region, and object localization is performed via appearance matching. 
    (b) The motion-centric paradigm takes concatenated point cloud frames as input and estimates relative motion based on segmented objects. 
    (c) Our proposed StreamTrack only takes the current frame as input, while historical features are fetched from a memory bank, allowing for the exploitation of multi-frame continuous motion for robust tracking.
    }
    \label{fig:teaser}
\end{figure}

\section{Introduction}
The rapid advancement of LiDAR technology has sparked a growing interest in point cloud-based vision solutions over recent years. 3D single object tracking (SOT) based on point clouds is a fundamental task that holds enormous potential for various applications, including autonomous driving and robotics. Nevertheless, 3D SOT remains a challenging and open problem owing to the inherent properties of point clouds, such as point sparsity, partial observation, and lack of texture information. The development of effective 3D SOT solutions continues to be an active research focus.

Most existing 3D SOT approaches~\cite{2019sc3d, 2020p2b, zheng2021box, hui20213dv2b, zhou2022pttr, hui20223dstnet} follow the prevalent paradigm of appearance matching (Fig.~\ref{fig:teaser}{(a)}), which originated from their 2D counterparts~\cite{tao2016siamese, guo2017learning, li2019siamrpn++}. Such matching-based methods perform feature matching between a cropped template and a search region to locate target objects but are prone to errors in cases of fast movement, occlusion, and misleading objects with similar appearances.
Recently, a new motion-centric paradigm~\cite{zheng2022beyond} (Fig.~\ref{fig:teaser}{(b})) has been proposed, which utilizes concatenated point clouds from two successive frames as input to preserve the motion connection. It performs motion estimation based on segmented foreground points to predict the relative motion and achieves outstanding tracking performance. However, it has two limitations. First, it neglects the dynamics (\textit{e.g.}, velocity and acceleration) contained in multi-frame historical movements, which could provide strong cues to future motion. Second, the segmentation stage eliminates background points, which might contain helpful contextual information for subsequent object localization. Erroneous segmentation could also affect tracking adversely.

Based on the aforementioned observations, we aim to propose a solution that can effectively exploit multi-frame continuous motion for accurate and robust object tracking. A straightforward method is to extend the existing motion-centric paradigm by concatenating points from multiple frames to form the input. However, this approach would result in high computational overhead and a potential problem that objects could exceed the predefined search range.
To address these issues, we propose a new framework for 3D SOT named \textit{StreamTrack}. As shown in Fig.~\ref{fig:teaser}{(c)}, we treat each tracking sequence as a \textit{stream}: at each timestamp, only the current frame is used as input, while historical features are stored in a live memory bank. Multi-frame features undergo a spatial-temporal relation modeling process to generate tracking predictions in an end-to-end manner. To achieve effective cross-frame message passing, we design a hybrid attention mechanism that can handle both long-range relation modeling and local geometric feature extraction simultaneously. To further improve the utilization of multi-frame features for robust tracking, we incorporate a contrastive sequence enhancement strategy where ground truth tracklets are used to augment training sequences and promote discrimination against false positives in a contrastive manner. This effectively mitigates the issue of target-switch, where a wrong object is tracked during tracking. We evaluate our proposed approach on KITTI, nuScenes, and Waymo datasets, and the experimental results demonstrate that StreamTrack achieves new state-of-the-art performance on all benchmarks.

The contributions of this work are summarized below: 
\textbf{1)} We identify an overlooked aspect in existing 3D SOT paradigms and propose StreamTrack -- a new paradigm that treats each tracking sequence as a stream and 
utilizes a memory bank for efficient exploitation of multi-frame continuous motion; 
\textbf{2)} We propose a hybrid attention mechanism that can handle both long-range relation modeling and local geometric feature extraction to achieve effective cross-frame message passing;
\textbf{3)} We design a contrastive sequence enhancement scheme that further improves the utilization of multi-frame features for robust tracking;
\textbf{4)} Experimental results on KITTI, nuScenes, and Waymo demonstrate that StreamTrack achieves new state-of-the-art performance while still being computationally efficient.

\begin{figure*}[t]
    \centering
    \includegraphics[width=1.0\linewidth]{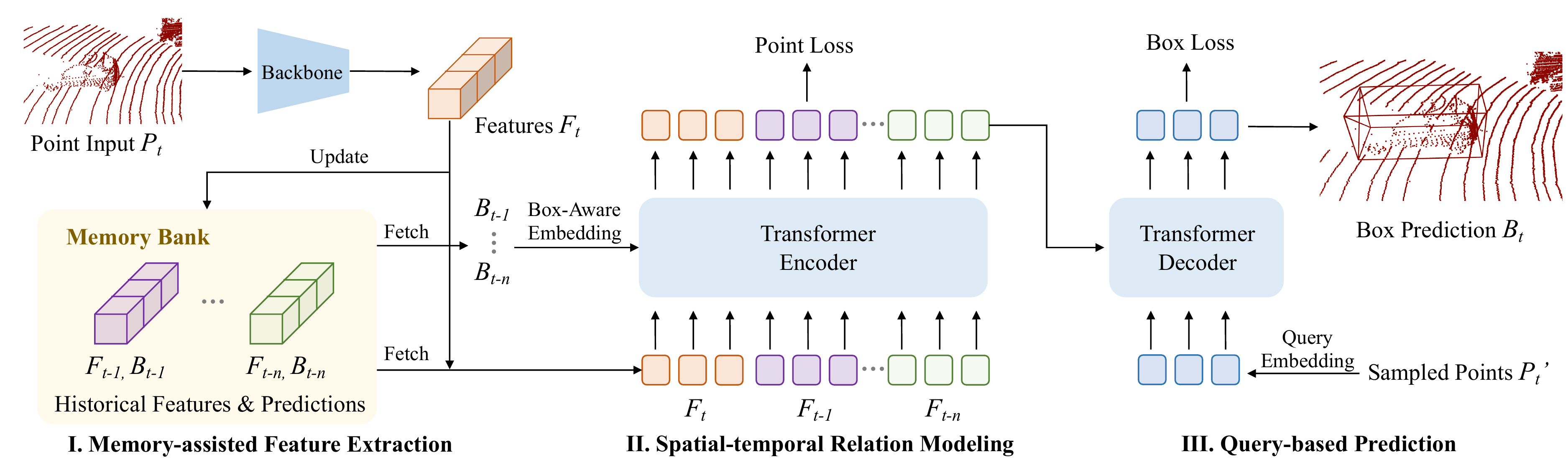}
    \caption{Overall architecture of StreamTrack. StreamTrack consists of three modules: memory-assisted feature extraction, spatial-temporal relation modeling, and query-based prediction. At timestamp $t$, StreamTrack only takes as input the current frame $P_t$, while historical features and box predictions are fetched from a memory bank for efficient computation. A Transformer encoder-decoder architecture is adopted for cross-frame message passing and the generation of tracking predictions.
    }
    \label{fig:model}
\end{figure*}

\section{Related Work}

\noindent\textbf{3D Single Object Tracking.} 
Given a point cloud sequence and the bounding box of an object in the first frame, the goal of 3D SOT is to locate the object in subsequent frames. The primary application of 3D SOT in autonomous driving is to enable a vehicle to follow a specific object (\textit{e.g.}, a car). 
Most existing 3D SOT approaches follow the matching-based paradigm, which is inspired by the success of Siamese networks in 2D tracking~\cite{tao2016siamese, li2018high, li2019siamrpn++}. As the pioneering work, SC3D~\cite{2019sc3d} generates a series of target proposals to match with the template based on feature similarities and selects the proposal with the top similarity. 
P2B~\cite{2020p2b} uses a Region Proposal Network~\cite{qi2019deep} for efficient proposal generation and employs Hough Voting to generate the tracking prediction. Motivated by the success of P2B, a series of follow-up work further improves the feature correlation operation or prediction generation with more sophisticated designs. For example, SA-P2B~\cite{zhou2021structure} designs an auxiliary task to learn the structure of objects. BAT~\cite{zheng2021box} encodes structural information with Box Cloud for individual points. MLVSNet~\cite{wang2021mlvsnet} enhances the feature aggregation with multi-level Hough Voting. V2B~\cite{hui20213dv2b} performs Voxel-to-BEV transformation for object localization on the densified feature maps. 
Inspired by the success of Transformer~\cite{vaswani2017attention} on computer vision tasks~\cite{liu2021swin, carion2020end}, several studies~\cite{zhou2022pttr, cui20213dlttr, shan2021ptt, hui20223dstnet, guo2022cmt, nie2023glt, xu2023cxtrack, luo2024exploring} incorporate Transformer for enhanced feature extraction and correlation modeling.

Despite its success, the matching-based paradigm breaks the motion connection between successive frames with template cropping and does not fully exploit the distortion-free property of point clouds. This makes it sensitive to distractors with similar geometric shapes~\cite{zheng2022beyond}. Recently, a motion-centric tracker M\textsuperscript{2}-Track~\cite{zheng2022beyond} achieves outstanding performance by tackling the tracking problem from the perspective of relative motion. Our proposed method also resorts to motion modeling but differs from M\textsuperscript{2}-Track by exploiting multi-frame continuous motion and having an end-to-end design.

\noindent\textbf{Contrastive Learning} works under the principle that similar sample pairs should be close in a learned embedding space, while distinct ones should be well separated. It has been extensively studied in representation learning~\cite{chen2020simple, he2020momentum, henaff2020data, oord2018representation, wu2018unsupervised, grill2020bootstrap} and achieved remarkable success in boosting the performance of downstream tasks. Several studies~\cite{lang2021contrastive, wu2022defenseidol, zhu2022conquer, yao2021g} extend contrastive learning to the supervised setting to learn more robust feature representations to reduce misclassification or wrong instance associations. Inspired by the above works, we design a contrastive sequence enhancement strategy to improve the robustness of tracking. To our best knowledge, this is the first effort to utilize contrastive learning in 3D SOT approaches.

\section{Methodology}
As illustrated in Fig.~\ref{fig:model}, StreamTrack consists of three modules: \textbf{1)} memory-assisted feature extraction, \textbf{2)} spatial-temporal relation modeling, and \textbf{3)} query-based prediction. We describe them in detail in the remainder of this section.

\subsection{Memory-assisted Feature Extraction} \label{sec:feature-extraction}
Existing 3D SOT methods 
typically rely on point clouds from two consecutive frames as input. However, multiple frames capture richer motion information, which can be exploited for more accurate and robust tracking. One possible approach to utilize multi-frame information is to concatenate points from a number of frames to form the input as in~\cite{zheng2022beyond}. However, such an approach becomes computationally expensive as the number of points increases. Additionally, fast-moving objects may go beyond the predefined search range, as a small range is typically used to reduce search complexity and improve efficiency. 

To address these issues, we propose a memory-assisted feature extraction scheme for the efficient utilization of multi-frame features. As shown in Fig.~\ref{fig:model}, given a point cloud frame $P_t \in \mathbb{R}^{N \times3}$ at timestamp $t$, where $N$ is the number of input points, we use a backbone model to extract the point features $F_t \in \mathbb{R}^{N' \times C}$, where $N'$ is the number of points sampled by the backbone.
We adopt PointNet++~\cite{qi2017pointnet++} as our backbone as it is widely used in existing 3D SOT methods~\cite{2020p2b, zheng2021box, zhou2022pttr}, although it is possible to use more sophisticated backbone networks to further improve the tracking performance. 
We employ a memory bank to store historical point features and box predictions of the past $n$ frames, which are denoted by $\{F_{t-i}\}_{i=1}^n$ and $\{B_{t-i}\}_{i=1}^n$, respectively. At the end of each iteration, we update the memory bank with $F_t$ and $B_t$ and discard those from the earliest frame ($F_{t-n}$ and $B_{t-n}$).

The memory bank design allows StreamTrack to bypass the repetitive computation of multi-frame point features and greatly improves the efficiency of our framework. Apart from only requiring the current frame as input at each timestamp, another major difference between our StreamTrack and M\textsuperscript{2}-Track is that M\textsuperscript{2}-Track utilizes a unified coordinate system for all input frames, whereas we shift the coordinate system for each frame to follow the movement of objects. Specifically, the canonical coordinate system defined by box prediction $B_{t-1}$ is employed for input $P_t$. 
This design allows reusing historical features without enlarging the input range to cover long-range movements while preserving the critical relative motion.

\subsection{Spatial-temporal Relation Modeling} \label{sec:relation}

The goal of this module is to model the cross-frame relation and propagate target information from past frames to the current frame for subsequent object localization. To deal with the complexity introduced by multi-frame features, we employ the attention mechanism of Transformer~\cite{vaswani2017attention} to leverage its 
strong capability to model long-range dependencies. Specifically, we use an encoder consisting of $L_{enc}$ stacked Transformer layers to encode point features and historical box locations. In the following text, we first describe the spatial-temporal relation modeling process using the original (vanilla) attention mechanism~\cite{vaswani2017attention} and then introduce our proposed hybrid attention designed for more effective cross-frame feature exchange.

\noindent\textbf{Vanilla Attention.}
For each Transformer layer, given point features $\{F_t,F_{t-1},...,$ $F_{t-n}\}$, we first concatenate them to form the input $F \in \mathbb{R}^{(n+1) \times N' \times C}$.
To incorporate the geometric locations of the points, 
we generate a position embedding \textit{PE}, which is of the same shape as $F$, by mapping the 3D coordinates of the points with an MLP. To distinguish points from different frames, we also add a learnable temporal embedding to \textit{PE} based on the temporal sequential order. Besides, the tracking process is conditioned on the prior knowledge of object locations in past frames. To incorporate past box locations,
we generate a point mask $M \in \mathbb{R}^{(n+1) \times N'}$ to indicate the objectiveness of each point as in~\cite{zheng2022beyond}. Concretely, $m_j^i \in M$ is defined as:
\begin{equation}
    m_j^i = \begin{cases}
            0 & \text{if} \; j \in [t-1, t-n] \; \text{and} \; p_j^i \; \text{is not in} \; B_j \\
            1 & \text{if} \; j \in [t-1, t-n] \; \text{and} \; p_j^i \; \text{is in} \; B_j \\
            0.5 & \text{if} \; j = t
            \end{cases}
\end{equation} 
where $i$ indexes the points and $j$ indexes the timestamps. Intuitively, $m_j^i$ can be viewed as the probability of point $p_j^i$ belonging to the foreground.
However, $M$ does not accurately encode the box location and orientation especially when points are sparse. \cite{zheng2021box} proposes to represent the point-to-box relation by including the distances from each point to the box center and 8 corners. We concatenate the said distances to $M$ and obtain the box-aware point mask $M' \in \mathbb{R}^{(n+1) \times N' \times (1+9)}$, where the distance values for the current frame are set to zero due to unknown box location. Similarly, we map $M'$ with an MLP to obtain a box-aware mask embedding \textit{ME}. Finally, we add the position embedding to $F$ to form query and key, while adding the mask embedding to $F$ to obtain value, and perform the attention computation as defined in~\cite{vaswani2017attention}:
\begin{align}
    Q = K &= F + \textit{PE} ; \;\; V = F + \textit{ME} \\
    F’ &= \text{Softmax}(\frac{QK^T}{\sqrt{d_k}})V
\end{align}
where $F'$ denotes the attention output and $d_k$ is the key dimension. A standard feedforward network (FFN) is further applied to generate the output of the Transformer layer.

\begin{figure}[t]
    \centering
    \includegraphics[width=0.75\linewidth]{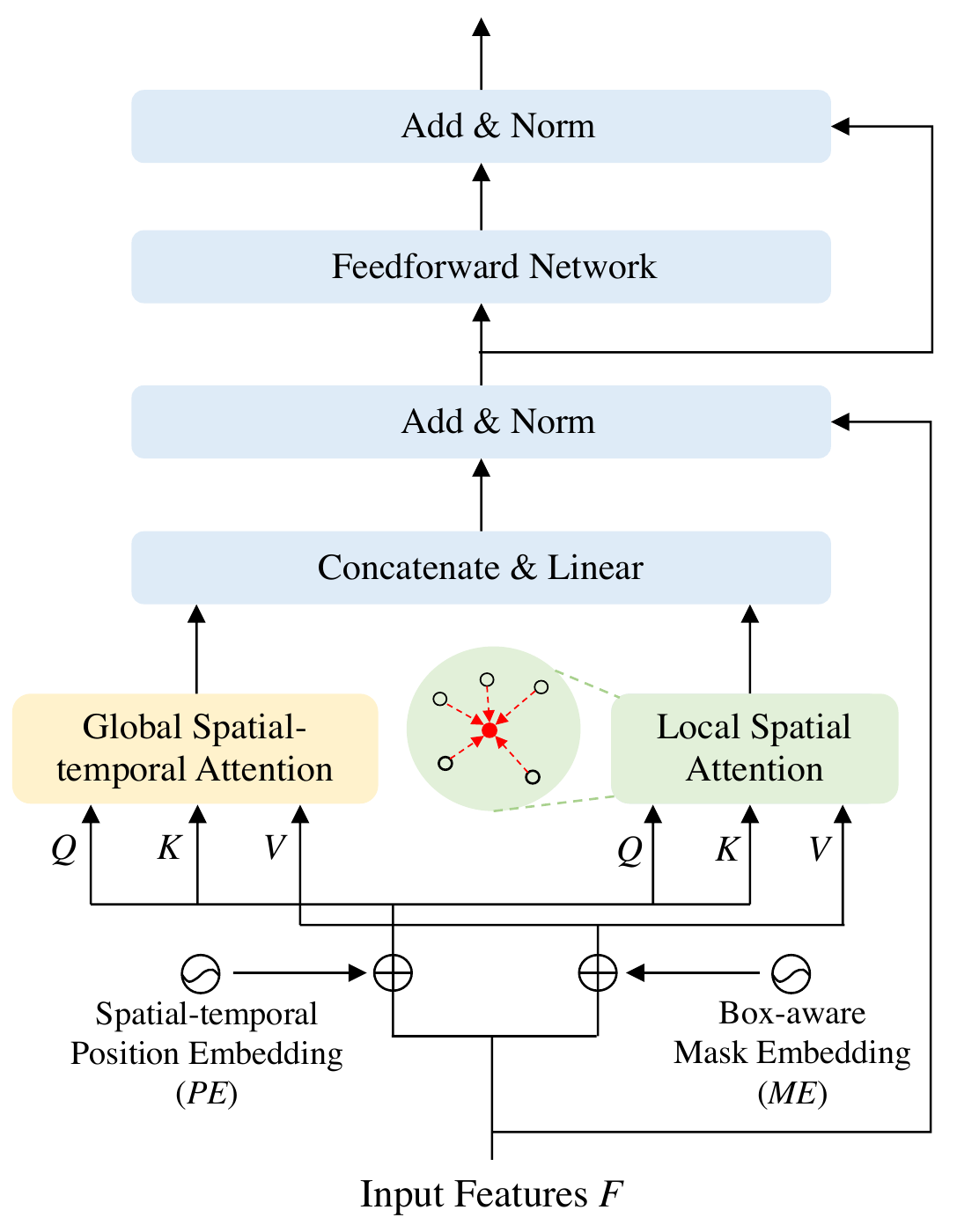}
    \caption{Architecture of the proposed hybrid attention. A local spatial attention module is introduced to work in parallel with global spatial-temporal attention to account for both local feature extraction and long-term relation modeling to achieve more effective cross-frame message passing.}
    \label{fig:hybrid}
\end{figure}

\noindent\textbf{Hybrid Attention.} 
Albeit the strong global relation modeling capability of Transformer, it treats each point equally without paying specific attention to local geometric structures, which have been proven to be crucial in various point representation learning studies~\cite{qi2017pointnet++, thomas2019kpconv, zhao2021point}. To achieve more effective cross-frame message passing, we design a hybrid attention mechanism that incorporates such inductive bias into Transformer. As shown in Fig.~\ref{fig:hybrid}, we introduce a local spatial attention operation in parallel with the regular global spatial-temporal attention to account for both local geometric feature extraction and long-range relation modeling. Specifically, we gather a local set of points $\{p_b| |p_b - p_a|<r\}$ for each input point $p_a$ with a predefined distance threshold $r$ and replace $K$ and $V$ in the vanilla attention with the corresponding local point features. 
The outputs of both attention modules are then concatenated and merged with a Linear layer. The proposed hybrid attention enhances the learning of local geometric structures to improve the modeling of spatial-temporal relations.

\noindent\textbf{Point Supervision.}
To promote cross-frame feature interaction and information propagation, we apply point-wise supervision on the encoder output. Specifically, for each encoder layer, we predict the point objectiveness $s$ and point-to-box distances $d$ as defined in the box-aware point mask generation process. The point loss $\mathcal{L}_{point}$ is formulated as:
\begin{equation}
    \mathcal{L}_{point} = \sum_{l}^{L_{enc}} (\lambda_s \mathcal{L}_{CE}(s_l, \hat{s}) + \lambda_d \mathcal{L}_{smooth-l1}(d_l, \hat{d}))
\end{equation}
where $\hat{(\cdot)}$ denotes the ground truth, $l$ indexes the encoder layers, and $\mathcal{L}_{CE}$ reprensents the cross-entropy loss.

\subsection{Query-based Prediction}  \label{sec:prediction}
Most existing 3D SOT methods~\cite{2020p2b, zheng2021box, zhou2022pttr, guo2022cmt, hui20223dstnet} employ point-based RPNs to generate tracking predictions. Inspired by DETR~\cite{carion2020enddetr}, we introduce a query-based prediction method in this work. As shown in Fig.~\ref{fig:model}, we first generate query embeddings based on the coordinates of the sampled points $P'_t$ using an MLP. The query embeddings are then input to a decoder consisting of $L_{dec}$ Transformer layers to interact with the encoder output for tracking prediction generation. 
Unlike RPN-based approaches that generate tracking predictions over the current frame, our approach allows each object query to interact with encoded features from all input frames.

\noindent\textbf{Box Supervision.}
For each decoder layer, a classification head and a regression head are applied to generate class predictions $c$ and box predictions $b$. Box matching is then performed to match the predictions to the ground truth box $\hat{b}$. The matching process is similar to the set-to-set matching in DETR except that we aim to find one single prediction that is best matched to the ground truth. We define the matching cost function with a semantic term and a geometric term:
\begin{equation}
    \mathcal{L}_{match} = -\lambda_{cls} c - \lambda_{giou} \mathcal{L}_{giou}(b, \hat{b})
\end{equation}
where $\mathcal{L}_{giou}$ measures the box overlap using GIoU~\cite{rezatofighi2019generalized}. We then select the prediction with the lowest matching cost as the positive prediction and set the classification target as one, while the rest are regarded as negative predictions with classification targets of zeros. Finally, the box loss can be formulated by:
\begin{equation}
    \mathcal{L}_{box} = \sum_{l}^{L_{dec}} (\lambda_{cls} \mathcal{L}_{focal}(c_l, \hat{c}_l) + \lambda_{reg} \mathcal{L}_{reg}(b_l^+, \hat{b}))
\end{equation}
where $\mathcal{L}_{focal}$ denotes the focal loss~\cite{lin2017focal}, and the regression loss $\mathcal{L}_{reg}$ consists of smooth-L1 and GIoU terms, which is detailed in the Appendix. The regression loss is only applied to the best-matched positive prediction $b_l^+$.

\begin{figure}[t]
    \centering
    \includegraphics[width=1.0\linewidth]{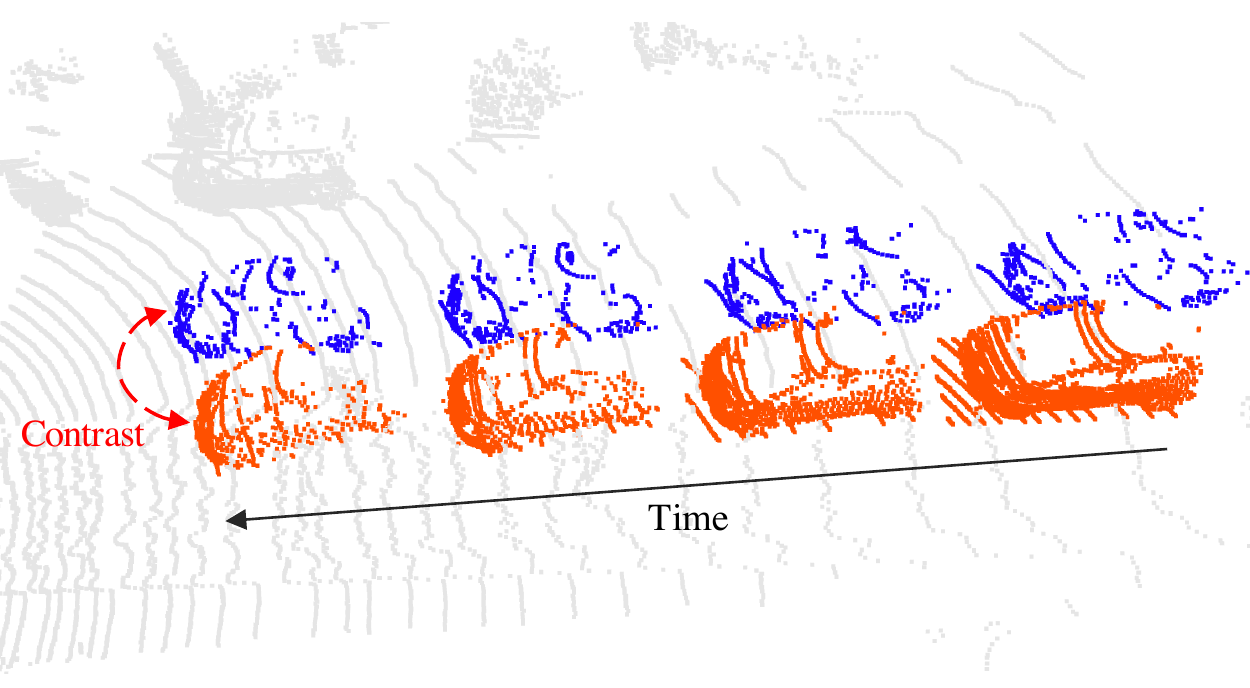}
    \caption{Illustration of contrastive sequence enhancement. Blue points denote the original target object in a tracking sequence, and orange points represent the added tracklet which serves as a negative sample. 
    An auxiliary contrastive loss is applied to further promote discrimination.
    }
    \label{fig:contrast}
\end{figure}

\noindent\textbf{Contrastive Sequence Enhancement.}
It has been identified that distractors (nearby objects with similar appearances) pose a significant challenge to 3D SOT as point clouds inherently provide limited appearance cues~\cite{zheng2022beyond}. We observe that there might not exist an abundance of distractors in the training sequences since typically only a small search region is considered, especially under data-constrained scenarios. As a result, tracking methods may not be fully trained to discriminate against negative targets. To this end, we propose a \textit{sequence enhancement} scheme by purposely attaching additional tracklets to training sequences to serve as negative samples.  As illustrated in Fig.~\ref{fig:contrast}, with a probability $\rho$, we randomly sample a tracklet of the same category from the training data and attach it to the input frames. The added tracklet is placed near the target tracklet with a random relative velocity to simulate object movements. Note that although the proposed sequence enhancement shares some similarities with the commonly used `GT-AUG'~\cite{yan2018second} in 3D detection, they differ in the following aspects. First, `GT-AUG' is usually applied to a single frame, while sequence enhancement works with sequential frames. Second, `GT-AUG' introduces additional positive targets, whereas sequence enhancement attaches negative samples to enhance discrimination.

Motivated by the success of contrastive learning~\cite{chen2020simple, he2020momentum, henaff2020data}, we introduce an auxiliary contrastive loss to explicitly enforce the separation between positive and negative targets in the feature embedding space. Specifically, we add an additional GT query as in~\cite{li2022dndetr} by generating its query embedding based on the center location of the ground truth bounding box. Note that the inclusion of the GT query does not have a significant impact on the tracking performance (see Appendix). The prediction generated by the GT query naturally forms a positive pair with the best-matched positive prediction and forms negative pairs with the remaining negative predictions. We follow the practice in MoCo~\cite{he2020momentum} to generate GT feature embedding $f_g$ with a momentum decoder obtained via exponential moving average, and the prediction embeddings are generated by projecting the decoder outputs with an MLP. The auxiliary contrastive loss is calculated using the InfoNCE~\cite{oord2018representation} loss over all decoder layers:
\begin{equation}
    \mathcal{L}_{aux} = \sum_{l}^{L_{dec}} -\text{log} \frac{\text{exp}(f_g^l \cdot f_{+}^l / \tau)}{\sum_{i=1}^{N'} \text{exp}(f_g^l \cdot f_i^l /\tau)}
\end{equation}
where $f_{+}^l$ denotes the feature embedding of the matched positive prediction, and $\tau$ is a temperature hyper-parameter~\cite{wu2018unsupervised}.
Note that contrastive sequence enhancement is only applied during training so that it brings no extra computation to the inference process.

\subsection{Training Loss}
We define the total loss function as the linear sum of the point loss, the box loss, and the auxiliary contrastive loss:
\begin{equation}
    \mathcal{L}_{total} = \lambda_{point} \mathcal{L}_{point} + \lambda_{box} \mathcal{L}_{box} + \lambda_{aux} \mathcal{L}_{aux}
\end{equation}
We include more details such as hyper-parameter values in the Appendix due to space constraints.

\begin{table*}[t]
\setlength{\tabcolsep}{20pt}
\begin{center}\setlength{\tabcolsep}{7pt}{
\small
\begin{tabular}{cc|cccccc|ccc}
\toprule[1.2pt]
 & Dataset & \multicolumn{6}{c|}{nuScense} & \multicolumn{3}{c}{Waymo} \\
 & Category & Car & Pedestrian &  Truck & Trailer & Bus & Mean & Vehicle & Pedestrian & Mean \\ 
 & Frame Count & 64,159 & 33,227 & 13,587 & 3,352 & 2,953 & 117,278 & 1,057,651 & 510,533 & 1,568,184 \\
\hline 
\multirow{5}{*}{\STAB{\rotatebox[origin=c]{90}{Success}}} &
SC3D & 22.31 & 11.29 & 30.67 & 35.28 & 29.35 & 20.70 & - & - & - \\
 & P2B  & 38.81 & 28.39 & 42.95 & 48.96 & 32.95 & 36.48 & 28.32 & 15.60 & 24.18 \\
 & BAT  & 40.73 & 28.83 & 45.34 & 52.59 & 35.44 & 38.10 & 35.62 & 22.05 & 31.20  \\
 & M\textsuperscript{2}-Track  & 55.85 & 32.10 & 57.36 & 57.61 & 51.39 & 49.23 & 43.62 & 42.10 & 43.13 \\
 & StreamTrack  & \textbf{62.05} & \textbf{38.43} & \textbf{64.67} & \textbf{66.67} & \textbf{60.66} & \textbf{55.75} & \textbf{60.23} & \textbf{47.07} & \textbf{55.95} \\

\hline
\multirow{5}{*}{\STAB{\rotatebox[origin=c]{90}{Precision}}}&
SC3D   & 21.93 & 12.65 & 27.73 & 28.12 & 24.08 & 20.20 & - & - & - \\
 & P2B  & 42.18 & 52.24 & 41.59 & 40.05 & 27.41 & 45.08 & 35.41 & 29.56 & 33.51 \\
 & BAT  & 43.29 & 53.32 & 42.58 & 44.89 & 28.01 & 45.71 & 44.15 & 36.79 & 41.75  \\
 & M\textsuperscript{2}-Track  & 65.09 & 60.92 & 59.54 & 58.26 & 51.44 & 62.73  & 61.64 & 67.31 & 63.48 \\
 & StreamTrack  & \textbf{70.81} & \textbf{68.58} & \textbf{66.60} & \textbf{64.27} & \textbf{59.74} & \textbf{69.22}  & 
 \textbf{72.61} & \textbf{70.44} & \textbf{71.90} \\
\bottomrule[1.2pt]
\end{tabular}
}
\end{center}
\caption{Performance comparison on nuScense and Waymo. \textit{Mean} performance is weighted by the number of frames.} \label{tab:nuscenes_waymo} 
\end{table*}

\section{Experiments}
\subsection{Experiment Setups}
\noindent\textbf{Datasets.}
We conduct extensive evaluations on three widely used datasets: KITTI~\cite{kitti}, nuScense~\cite{caesar2020nuscenes}, and Waymo Open Dataset~\cite{waymo}. 
For KITTI, we follow the data split defined in~\cite{2019sc3d}.
The nuScenes and Waymo datasets are of significantly larger scales as compared to KITTI.
We follow the implementation of~\cite{zheng2022beyond} for these two datasets, except we randomly sample 10\% of the tracklets for training on the Waymo dataset due to its overwhelming sample size. Testing is conducted over all test samples to ensure fair comparisons.

\noindent\textbf{Evaluation Metrics.}
We follow existing studies~\cite{2019sc3d, 2020p2b} and use the One Pass Evaluation~\cite{kristan2016novel} to measure the \textit{Success} and \textit{Precision} of tracking predictions. \textit{Success} is computed from the intersection over union (IOU) of the predicted bounding box and the ground truth box, while \textit{Precision} is defined as the area under the curve (AUC) for the distance between two box centers from 0 to 2 meters. 

\begin{table}[t]
\begin{center}\setlength{\tabcolsep}{1.5pt}{
\small
\begin{tabular}{c|ccccc}
\toprule[1.2pt]
 Category & Car & Pedestrian &  Van & Cyclist &  Mean \\
 Frame Count & 6424 & 6088 & 1248 & 308 & 14068 \\
\hline
SC3D  & 41.3/57.9 & 18.2/37.8 & 40.4/47.0 & 41.5/70.4 & 31.2/48.5 \\
P2B  & 56.2/72.8 & 28.7/49.6 & 40.8/48.4 & 32.1/44.7 & 42.4/60.0 \\
MLVSNet  & 56.0/74.0 & 34.1/61.1 & 52.0/61.4 & 34.3/44.5 & 45.7/66.7 \\
BAT  & 65.4/78.9 & 45.7/74.5 & 52.4/67.0 & 33.7/45.4 & 55.0/75.2 \\
PTTR  & 65.2/77.4 & 50.9/81.6  & 52.5/61.8 & 65.1/90.5 & 57.9/78.1 \\
V2B  & 70.5/81.3 & 48.3/73.5 & 50.1/58.0 & 40.8/49.7 & 58.4/75.2 \\
CMT  & 70.5/81.9 & 49.1/75.5 & 54.1/64.1 & 55.1/82.4 & 59.4/77.6 \\
GLT-T  & 68.2/82.1 & 52.4/78.8 & 52.6/62.9 & 68.9/92.1 & 60.1/79.3 \\
STNet  & 72.1/\textbf{84.0} & 49.9/77.2 & 58.0/70.6 & 73.5/93.7 & 61.3/80.1 \\
M\textsuperscript{2}-Track  & 65.5/80.8 & 61.5/88.2 & 53.8/70.7 & 73.2/93.5 & 62.9/83.4 \\
CXTrack  & 69.1/81.6 & 67.0/91.5 & 60.0/71.8 & 74.2/94.3 & 67.5/85.3 \\
StreamTrack  & \textbf{72.6}/83.7 & \textbf{70.5}/\textbf{94.7} & \textbf{61.0}/\textbf{76.9} & \textbf{78.1}/\textbf{94.6} & \textbf{70.8}/\textbf{88.1} \\
\bottomrule[1.2pt]
\end{tabular}
}
\end{center}
\caption{Performance comparison on the KITTI dataset. Success / Precision are reported.
} 
\label{tab:kitti_experiment} 
\end {table}

\subsection{Benchmarking Results}
\noindent\textbf{Results on KITTI.} 
On the KITTI dataset, we compare with state-of-the-art methods SC3D~\cite{2019sc3d}, P2B~\cite{2020p2b}, MLVSNet~\cite{wang2021mlvsnet}, BAT~\cite{zheng2021box}, PTTR~\cite{zhou2022pttr}, V2B~\cite{hui20213dv2b}, CMT~\cite{zheng2022beyond}, and CXTrack~\cite{xu2023cxtrack}.
As shown in Tab.~\ref{tab:kitti_experiment}, the proposed StreamTrack outperforms existing methods by notable margins in terms of average success and precision, while achieving top performance for most categories. Notably, matching-based methods (\textit{e.g.}, STNet~\cite{hui20223dstnet} and CMT~\cite{guo2022cmt}) tend to perform well on Car and Van, while motion-centric method M\textsuperscript{2}-Track is competitive on Pedestrian and Cyclist. We conjecture that cars and vans are rigid in shape and relatively sizeable, which makes them suitable for appearance matching. In contrast, humans are non-rigid and often appear in crowds, which poses challenges to the matching process. On the other hand, although M\textsuperscript{2}-Track is more robust to distractors due to its motion-centric property, its use of simple operations (\textit{e.g.}, MLP and max-pooling) for feature extraction limits the capability of learning geometric structures. Our proposed StreamTrack employs hybrid attention for effective geometric feature extraction and leverages multi-frame continuous motion for robust tracking, thus achieving balanced performance.

\noindent\textbf{Results on nuScenes and Waymo.} 
We compare StreamTrack with methods evaluated under the same setting on nuScenes and Waymo. As shown in Tab.~\ref{tab:nuscenes_waymo}, StreamTrack achieves new state-of-the-art performance for all categories and outperforms compared methods by clear margins. On the Waymo dataset, despite using only 10\% of the training samples, our proposed method still outperforms existing methods trained with the full train set. The nuScenes dataset is known for its low point density as it is collected using 32-beam LiDARs as compared to 64-beam for other datasets, while the Waymo dataset captures complex traffic scenes with numerous objects. The outstanding performance of StreamTrack on both datasets demonstrates its strong capability of tracking objects under challenging scenarios.

\noindent\textbf{Inference Speed.}
StreamTrack achieves an inference speed of 40.7 FPS when running on a single NVIDIA V100 GPU, which is on par with existing matching-based methods (\textit{e.g.}, P2B and BAT). Please refer to the Appendix for more information.

\begin{table*}[t]
\begin{center}\setlength{\tabcolsep}{5pt}{
\small
\begin{tabular}{c|ccccc|ccccc}
\toprule[1.2pt]
\# & Hybrid & Point Sup & Query Pred & Seq Enhance & Contrast & Car & Pedestrian &  Van & Cyclist &  Mean \\
\midrule
  1 &  & \cmark &   \cmark &  \cmark &  \cmark & 70.5 / 81.2 &	69.1 / 94.3 &	57.4 / 70.7 &	76.5 / 93.4 & 68.9 / 86.2 \\
 2 & \cmark &  &  \cmark &  \cmark &  \cmark &  70.4 / 81.0 & 62.4 / 90.7	 & 58.7 / 72.3	 & 72.2 / 93.6  & 65.9 / 84.7\\
 3 & \cmark  & \cmark  &   &  \cmark &  & 70.1 / 81.0 &	65.1 / 92.3 &	59.0 / 74.3 &	75.4 / 94.0 &	67.1 / 85.6 \\ 
 4 & \cmark & \cmark &   \cmark &   &   & 70.1 / 80.6 &  65.2 / 90.6	 & 52.1 / 63.0	 & 75.1 / 94.3	 & 66.5 / 83.7 \\
5 & \cmark & \cmark &   \cmark & \cmark  &  & 70.6 / 81.3 &	67.8 / 93.1 &	57.4 / 69.0 &	78.6 / 95.1 & 68.4 / 85.6 \\
  6 & \cmark & \cmark &   \cmark &  & \cmark  & 70.9 / 81.2 &	67.6 / 93.5 &	53.2 / 63.5 &	77.2 / 94.7 & 68.0 / 85.2 \\
7 & \cmark & \cmark &   \cmark &  \cmark &  \cmark & 72.6 / 83.7 &	70.5 / 94.7 &	61.0 / 76.9 &	78.1 / 94.6 & 70.8 / 88.1 \\
\bottomrule[1.2pt]
\end{tabular}
}
\end{center}
\caption{Ablation studies on model components. `Hybrid' denotes hybrid attention. `Point Sup' denotes point supervision. `Query Pred' denotes our proposed query-based prediction paradigm. When `Query Pred' is disabled, we replace the decoder with the prediction head in~\cite{zheng2022beyond}. `Seq Enhance' denotes sequence enhancement, and `Contrast' denotes auxiliary contrastive loss. Success / Precision are reported.}
\label{tab:model_components} 
\vspace{-1mm}
\end{table*}

\begin{figure*}[t]
    \centering
    \includegraphics[width=1.0\linewidth]{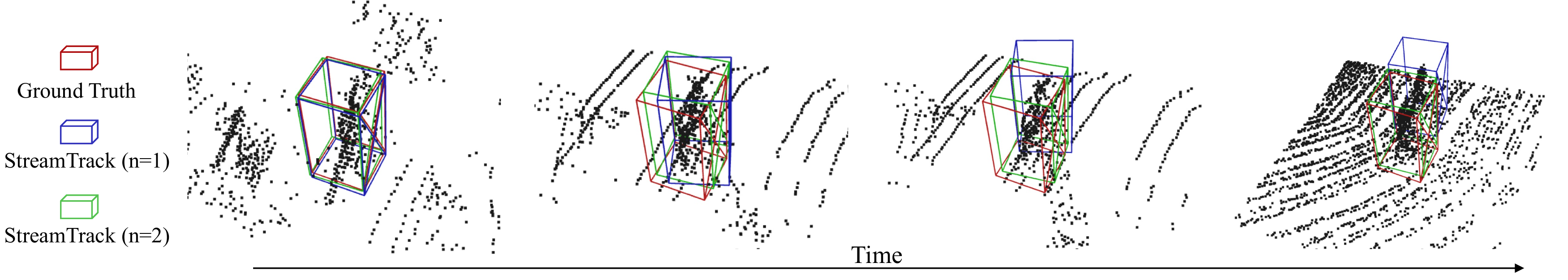}
    \caption{Visualization of tracking predictions on a Pedestrian sequence in which distractors exist. When $n=1$, StreamTrack only relies on one historical frame, which is similar to the existing motion-centric paradigm~\cite{zheng2022beyond}. It demonstrates that the exploitation of multi-frame continuous motion improves the tracking robustness effectively.}
    \label{fig:visual}
\end{figure*}

\subsection{Ablation Study} \label{sec:analysis}

\begin{figure}[t]
    \centering
    \includegraphics[width=0.9\linewidth]{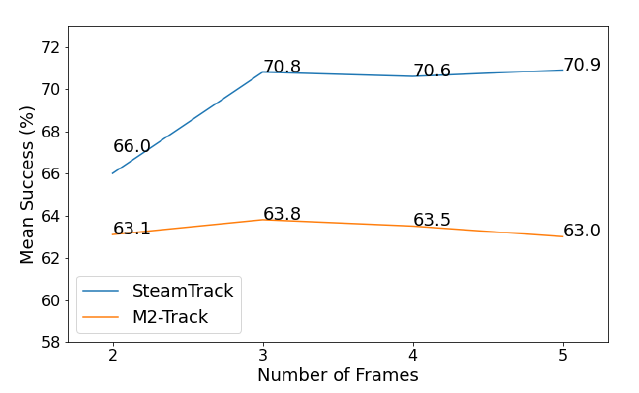}
    \caption{Mean Success vs Number of Frames that are used for predictions, evaluated over the KITTI dataset.}
    \label{fig:frame}
\end{figure}

\noindent\textbf{Effectiveness of Continuous Motion Modeling.} 
The key motivation of StreamTrack is to exploit multi-frame continuous motion for more informed and robust tracking. We conduct experiments to study the impact on performance when information from different numbers of frames (including $n$ historical frames and the current frame) is used to generate the tracking prediction. For a more comprehensive evaluation, we also extend M\textsuperscript{2}-Track~\cite{zheng2022beyond} to the multi-frame setting by concatenating points from multiple frames to form its input. 
 As shown in Fig.~\ref{fig:frame}, a significant improvement of 4.8\% in success is observed for StreamTrack when the number of frames increases from 2 to 3, and the performance stabilizes when the frame number is further increased. We hypothesize that two historical frames could already provide strong motion cues (\textit{e.g.}, velocity and acceleration) to aid the tracking process, while further increasing the frame count leads to marginal gains and extra complexity. In Fig.~\ref{fig:visual}, we visualize a tracking sequence to further demonstrate the effectiveness of exploiting multi-frame information in improving the robustness of tracking. Based on the experimental results, we use two historical frames ($n=2$) in our default setting considering the efficiency aspect. 
 On the other hand, the performance of M\textsuperscript{2}-Track only improves marginally when multiple frames are used and a downward trend is observed when the number of frames further increases. This could be attributed to the simple architecture of M\textsuperscript{2}-Track, which is not designed to handle the complex multi-frame feature interaction. 

\noindent\textbf{Effectiveness of Model Components.}
We conduct comprehensive ablation studies to investigate the effectiveness of the building components of StreamTrack. As shown in Tab.~\ref{tab:model_components}, removing hybrid attention (comparing $\#$1 and $\#7$) leads to a decrease of 1.9\% in mean success and precision, which validates that the local attention design complements the vanilla global attention mechanism. $\#$2 studies the impact of point supervision, and we observe a notable decrease of 4.9\% and 3.4\% in mean success and precision compared to our default setting ($\#7$) when the point loss on encoder predictions is removed. In particular, the Pedestrian and Cyclist categories suffer from larger drops, which implies that low-level supervision is more important for objects of smaller sizes. 
To investigate the effectiveness of the proposed query-based prediction, we replace the decoder of StreamTrack with the prediction head in M\textsuperscript{2}-Track. Note that the contrastive loss is dependent on the query-based design so it is not included. By comparing $\#3$ and $\#5$, it can be seen that our query-based prediction achieves improved overall performance, which shows the advantage of utilizing global multi-frame features for prediction generation in our design. 
From $\#$4 to $\#$7, it can be observed that both sequence enhancement and the auxiliary contrastive loss have a positive impact on the tracking performance. Moreover, they appear to be complementary to each other, as the performance is further improved when both are applied. 
Remarkably, sequence enhancement introduces a substantial performance improvement (+5.3\% in success) to the Van category, which has a limited number of training samples. This shows the potential of the sequence enhancement technique under data-constrained settings.

\section{Conclusion}
This paper presents StreamTrack, a new framework for 3D single object tracking (SOT) that considers each tracking sequence as a continuous stream and leverages the multi-frame motion information for more robust tracking. Our approach employs a memory-based feature extraction method to efficiently utilize multi-frame features and introduces hybrid attention to model spatial-temporal relations more effectively. We have also proposed a contrastive sequence enhancement strategy to improve the utilization of sequential information for false positive reduction. Our experimental results demonstrate that StreamTrack achieves state-of-the-art performance. We hope our work can serve as a baseline for utilizing sequential information in 3D SOT and inspire future research in this field.

\section*{Acknowledgements} 
This study is supported under the RIE2020 Industry Alignment Fund – Industry Collaboration Projects (IAF-ICP) Funding Initiative, as well as cash and in-kind contribution from the industry partner(s).

\bibliography{aaai24}

\newpage
\section{Appendix}
In this appendix, we provide additional discussions, implementation details, experimental results, and visualizations, which are not included in the main paper due to space limitations.

\subsection{Discussion}
\noindent\textbf{Relationship to 2D Tracking Approaches Utilizing Multi-frame Information.} 
A number of 2D tracking methods \cite{wang2021transformer, fu2021stmtrack, cai2022memot, xie2023videotrack} have explored leveraging multi-frame information to improve model performance. The fundamental difference between these methods and our proposed StreamTrack is that they essentially rely on appearance information for cross-frame association as in the matching-based paradigm, while we leverage the unique distortion-free property of point clouds and model the continuous motion. We achieve this by preserving the relative motion in our framework and incorporating the geometric positions of points and bounding boxes in the spatial-temporal modeling process, as introduced in the paper. Besides, we have also introduced new model components such as hybrid attention and contrastive sequence enhancement to further improve the effectiveness of motion modeling for robust tracking.

\subsection{Implementation Details}

\noindent\textbf{Loss Function.}
Our box prediction $b \in \mathbb{R}^{N' \times 4}$ consists of a 3-dimensional offset prediction $o \in \mathbb{R}^{N' \times 3}$ as well as a rotation prediction $\theta \in \mathbb{R}^{N'}$. The regression loss is defined as:
\begin{align*}
    \mathcal{L}_{reg} = &
    \lambda_{offset} \mathcal{L}_{smooth-l1}(o^+, \hat{o}) \\
+ & \lambda_{\theta} \mathcal{L}_{smooth-l1}(\theta^+, \hat{\theta}) 
+ \lambda_{giou} \mathcal{L}_{giou}(b^+, \hat{b})
\end{align*}
where $\hat{(\cdot)}$ denotes the ground truth and $(\cdot)^+$ denotes the prediction of the best-matched positive sample, on which the regression loss is computed. The loss coefficients are empirically set as $\lambda_{offset}=1.0$, $\lambda_{\theta}=5$, $\lambda_{giou}=0.25$, respectively. The rest of the coefficients are set as follows: $\lambda_{CE}=0.1$, $\lambda_{d}=1.0$, $\lambda_{cls}=0.1$, $\lambda_{reg}=1.0$, $\lambda_{point}=1.0$, $\lambda_{box}=2.0$, $\lambda_{aux}=0.05$.

\begin{figure}[t]
    \centering
    \includegraphics[width=0.9\linewidth]{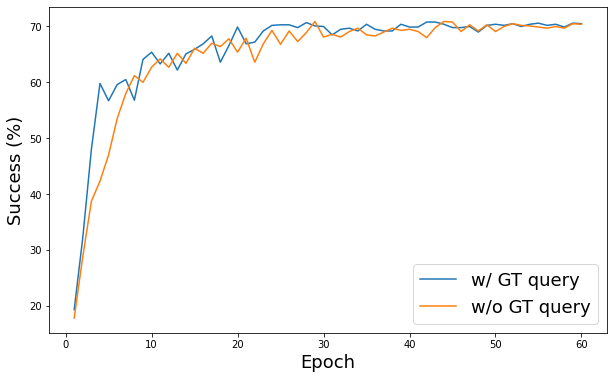}
    \caption{\textit{Success} vs \textit{Epoch} of the Car category on the KITTI dataset. The GT query speeds up the convergence in the early stage of training but does not have a significant impact on the tracking performance.}
    \label{fig:convergence}
\end{figure}

\smallskip
\noindent\textbf{Network.}
At each timestamp, we follow M\textsuperscript{2}-Track \cite{zheng2022beyond} to randomly sample $N=1024$ points to form the input. Our backbone network PointNet++ \cite{qi2017pointnet++} consists of two set abstraction layers with ball query radii 0.3 and 0.5 meters, which downsample the input to 512 and 128 points, respectively. The Transformer encoder consists of $L_{enc}=3$ encoder layers, and we set the distance threshold $r$ for hybrid attention as [0.6, 1.0, 1.5]. The Transformer decoder also contains $L_{dec}=3$ layers, and the number of object queries is equal to the number of points sampled by the backbone ($N'=128$). For contrastive sequence enhancement, we use a probability of $\rho=0.3$ to attach sampled tracklets to training sequences. We adopt a momentum value \cite{he2020momentum} of 0.99 for the momentum decoder used for GT feature embedding generation, and the temperature $\tau$ is set as 1.0.

\smallskip
\noindent\textbf{Data Augmentation.}
Similar to M\textsuperscript{2}-Track~\cite{zheng2022beyond}, we apply box augmentation and motion augmentation to the input data during training. For box augmentation, we follow existing methods~\cite{2020p2b, zheng2021box, zhou2022pttr, zheng2022beyond} to add a small random shift to bounding boxes on historical frames to simulate prediction errors during inference. For motion augmentation, M\textsuperscript{2}-Track employs object-wise augmentation by randomly transforming bounding boxes together with the points inside to diversify the relative motion. We instead perform frame-wise augmentation as in matching-based methods~\cite{2020p2b, zheng2021box, zhou2022pttr} by transforming all points from individual frames since our prediction process is based on global point features rather than only foreground points as in M\textsuperscript{2}-Track.

\smallskip
\noindent\textbf{Training and Inference.}
During training, we randomly sample $n+1$ consecutive frames from a tracking sequence to form a training sample, and the point features of all frames are extracted by the backbone network on the fly.
During inference, we store historical features and box predictions in the memory bank for efficient computation.

On the KITTI dataset, we train our model for 60 epochs with a batch size of 64. Adam \cite{kingma2014adam} optimizer is employed for model optimization with an initial learning rate of 0.0003. The learning rate is decayed by a factor of 10 every 25 epochs. On nuScense, we train our model for 30 epochs with a batch size of 128 on the Car and Pedestrian categories due to their large sample sizes. For the rest of the categories, we use the same setting as KITTI.
On Waymo, our model is trained for 10 epochs with a batch size of 256.
All the experiments are conducted using Pytorch 1.7 and NVIDIA V100 GPUS.

\begin{figure*}[t]
    \centering
    \includegraphics[width=1.0\linewidth]{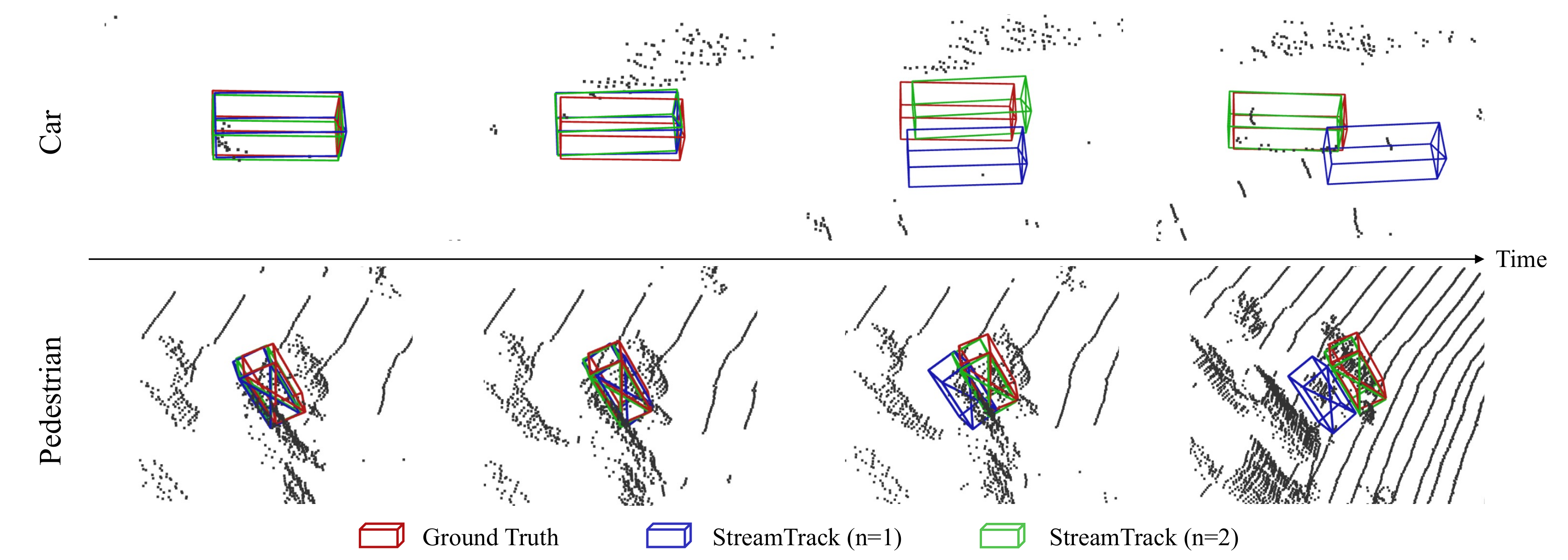}
    \caption{Visualization of the tracking predictions of our proposed StreamTrack. By leveraging multi-frame continuous motion (when the number of historical frames $n>1$), StreamTrack achieves more robust tracking results.}
    \label{fig:v2}
\end{figure*}

\subsection{Additional Experimental Results}
\noindent\textbf{Influence of GT Query.}
In our proposed contrastive sequence enhancement, we introduce an extra GT query as in \cite{li2022dndetr} to function as an `anchor' to form positive and negative pairs with the box predictions. As shown in Fig.~\ref{fig:convergence}, the GT query does not have a significant impact on the tracking performance although it speeds up the model convergence at the early stage of training. It validates that the performance gain is brought by the auxiliary contrastive loss instead of the GT query.

\begin{table}[t]
\centering
\setlength{\tabcolsep}{8pt}
\scalebox{0.9}{
\begin{tabular}{c|ccccc}
\toprule
$L_{enc}$ & Car & Pedestrian &  Van & Cyclist &  Mean \\
\hline \hline
1  & 70.3	 & 65.8	 & 59.4	 & 75.4	 & 67.5 \\
2  & 70.9	 & 68.2	 & 59.2	 & 76.8	 & 68.8 \\
3  & 72.6	 & 70.5	 & 61.0	 & 78.1   & 70.8\\
4  & 72.5	 & 70.3	 & 61.7	 & 77.8	 & 70.7 \\
\bottomrule
\end{tabular}
}
\centering
\caption{Influence of the number of encoder layers $L_{enc}$ when $L_{dec}$ is fixed at 3. The \textit{success} metric is reported.}
\label{tab:encoder}

\hfill

\centering
\setlength{\tabcolsep}{8pt}
\scalebox{0.9}{
\begin{tabular}{c|ccccc}
\toprule
$L_{dec}$ & Car & Pedestrian &  Van & Cyclist &  Mean \\
\hline \hline
1   & 69.8	 & 69.2	 & 60.5	 & 75.2	 & 68.8 \\
2   & 70.9	 & 69.8	 & 60.8	 & 75.5	 & 69.6 \\
3  & 72.6	 & 70.5	 & 61.0	 & 78.1   & 70.8 \\
4   & 71.3	 & 67.2	 & 58.6	 & 77.2	 & 68.5 \\
\bottomrule
\end{tabular}
}
\centering
\caption{Influence of the number of decoder layers $L_{dec}$ when $L_{enc}$ is fixed at 3. The \textit{success} metric is reported.}
\label{tab:decoder}
\end{table}

\smallskip
\noindent\textbf{Influence of the Number of Transformer Layers.}
Tab.~\ref{tab:encoder} and Tab.~\ref{tab:decoder} report the performance of our proposed StreamTrack when different numbers of encoder and decoder layers are used. It can be observed that the model achieves the best performance when both the encoder and decoder have 3 layers. Notably, StreamTrack still generates competitive results when fewer transformer layers are employed, which offers good flexibility under resource-constrained settings. 

\smallskip
\noindent\textbf{Influence speed.}
Tab.~\ref{tab:speed} presents the average inference speed, in the form of frame per second (FPS), measured using all test frames from the Car category on the KITTI dataset. 
We compare our StreamTrack with a number of representative methods, and the speed is measured using the same platform with a single NVIDIA V100 GPU. Thanks to the memory bank design, our proposed StreamTrack achieves a running speed of 40.7 FPS despite the relation modeling based on multi-frame features, which is on par with existing matching-based methods (\textit{e.g.}, P2B and BAT). Notably, M\textsuperscript{2}-Track is exceptionally fast due to its simple architecture which only consists of pooling operations and MLPs. However, such a design is inadequate for the effective exploration of multi-frame information (see Fig.~\ref{fig:frame}).

\begin{table}[t]
\setlength{\tabcolsep}{24pt}
\small
\begin{center}\setlength{\tabcolsep}{6pt}{
\scalebox{1.0}{
\begin{tabular}{c|ccc}
\toprule[1.2pt]
 Method & FPS & Success & Precision \\
\hline
P2B (CVPR'20) & 38.8 & 42.4 & 60.0 \\
BAT (ICCV'21)  & 40.9 & 55.0 & 75.2 \\
STNet (ECCV'22) & 30.1 & 61.3 & 80.1 \\
M\textsuperscript{2}-Track (CVPR'22) & \textbf{54.8} & 62.9 & 83.4 \\
StreamTrack (Ours) & 40.7 & \textbf{70.8} & \textbf{88.1} \\
\bottomrule[1.2pt]
\end {tabular}}
}
\end{center}
\caption{Inference speed comparison on the KITTI dataset.} 
\label{tab:speed} 
\end {table}

\subsection{Additional Visualizations}
In Fig.~\ref{fig:v2}, we provide more visualizations to demonstrate 
the effectiveness of modeling continuous motion for robust tracking with point clouds.

\end{document}